%% file: 00_main.tex
\documentclass[letterpaper, 10pt, conference]{ieeeconf}
\IEEEoverridecommandlockouts
\usepackage{mathtools}
\usepackage{cite}
\usepackage[dvipsnames]{xcolor}
\usepackage{amsmath,amssymb,amsfonts}
\usepackage[font=small, labelfont=bf]{caption}
\usepackage[ruled, lined, linesnumbered, commentsnumbered, longend, noend]{algorithm2e}
\usepackage{hyperref}
\usepackage{algorithmic}
\usepackage{graphicx}
\usepackage{textcomp}
\usepackage{breqn}
\usepackage{gensymb}
\usepackage{cuted}
\usepackage{capt-of}
\usepackage{booktabs}
\usepackage{array}
\usepackage{diagbox}
\usepackage{pifont}
\usepackage{multirow}
\usepackage{makecell}
\usepackage{bm}
\usepackage{siunitx}
\usepackage{gradient-text}
\usepackage{colortbl}

\usepackage{pifont}
\newcommand{\cmark}{\ding{51}}
\newcommand{\xmark}{\textcolor{gray}{\ding{55}}}

\usepackage{fancyhdr} 
\def\BibTeX{{\rm B\kern-.05em{\sc i\kern-.025em b}\kern-.08em
    T\kern-.1667em\lower.7ex\hbox{E}\kern-.125emX}}

\definecolor{pink}{RGB}{255, 192, 203}
\definecolor{lightblue}{HTML}{77AADD}
\definecolor{lightorange}{HTML}{EE8866}
\definecolor{violet}{HTML}{C299DD}
\definecolor{tealgreen}{HTML}{44BB99}
\definecolor{lightyellow}{HTML}{D0B866}
\definecolor{apricot}{HTML}{DD9977}
\usepackage{siunitx}

\definecolor{pink}{RGB}{255, 192, 203}

\definecolor{darkgreen}{RGB}{83, 199, 34}
\newcommand{\boldgreen}[1]{\textcolor{darkgreen}{\textbf{#1}}}
\newcommand{\boldred}[1]{\textcolor{BrickRed}{\textbf{#1}}}
\newcommand{\start}{\textcolor{gray!110}{\ding{108}}}
\newcommand{\goal}{\textcolor{gray!110}{\ding{54}}}

\title{\LARGE \bf $\mathrm{\textbf{X}}$\textit{FlowMP}: Task-Conditioned Motion Fields for Generative Robot Planning with Schr{\"o}dinger Bridges}

\author{\quad Khang Nguyen$^{1}$ and Minh Nhat Vu$^{2,3}$ 
\thanks{$^{1}$Mohamed bin Zayed University of Artificial Intelligence (MBZUAI), Abu Dhabi, UAE}
\thanks{$^{2}$Automation \& Control Institute (ACIN), TU Wien, Vienna, Austria} 
\thanks{$^{3}$Austrian Institute of Technology (AIT) GmbH, Vienna, Austria}
\thanks{$^{\dagger}$Corresponding authors / e-mails: \texttt{\footnotesize khang.nguyen@mbzuai.ac.ae}, \texttt{\footnotesize minh.vu@tuwien.ac.at}}
}

\begin{document}

\maketitle
\thispagestyle{empty}
\pagestyle{empty}

\begin{abstract}
    Generative robotic motion planning requires not only the synthesis of smooth and collision-free trajectories but also feasibility across diverse tasks and dynamic constraints. Prior planning methods, both traditional and generative, often struggle to incorporate high-level semantics with low-level constraints, especially the nexus between task configurations and motion controllability. In this work, we present $\mathbf{X}$\textit{FlowMP}, a task-conditioned generative motion planner that models robot trajectory evolution as entropic flows bridging stochastic noises and expert demonstrations via Schr{\"o}dinger bridges given the inquiry task configuration. In specific, our method leverages Schr{\"o}dinger bridges as a conditional flow matching coupled with a score function to learn motion fields with high-order dynamics while encoding start-goal configurations, enabling the generation of collision-free and dynamically-feasible motions. Through evaluations, $\mathbf{X}$\textit{FlowMP} achieves up to $53.79\%$ lower maximum mean discrepancy, $36.36\%$ smoother motions, and $39.88\%$ lower energy consumption while comparing to the next-best baseline on the \texttt{RobotPointMass} benchmark and also reducing short-horizon planning time by $11.72\%$. On long-horizon motions in the \texttt{LASA Handwriting} dataset, our method maintains the trajectories with $1.26\%$ lower maximum mean discrepancy, $3.96\%$ smoother, and $31.97\%$ lower energy. We further demonstrate the practicality of our method on the Kinova Gen3 manipulator, executing planning motions and confirming its robustness in real-world settings.
\end{abstract}

    

\input{01_introduction}
\input{02_related_work}
\input{03_methodology}
\input{04_evaluation}
\input{05_conclusions}

\bibliographystyle{IEEEtran}
\bibliography{IEEEabrv, references}

\end{document}

%% file: 01_introduction.tex
\section{Introduction}
Task-conditioned motion planning is one of the central problems in robot planning, entailing the systems to generate trajectories that satisfy specific constraints defined by the task at hand. Traditional approaches often rely on optimization-based methods, including sampling-based planners \cite{kavraki2002probabilistic}, graph search algorithms \cite{lavalle2001rapidly}, and trajectory optimization techniques \cite{ratliff2009chomp, schulman2013finding}. As task requirements are encoded through cost functions that find optimal paths through search or iterative refinement, such approaches often struggle with scalability, adaptability, and generalization to diverse, complex tasks \cite{papadopoulos2014analysis, beeching2020learning}. Moreover, these methods are computationally expensive when computing trajectories from scratch and heavily rely on good initialization. In the meantime, generative motion planning offers an alternative approach that directly learns the distribution of expert trajectories from data, enabling the fast sampling of diverse, context-adaptive solutions that naturally balance task semantics with physical feasibility.

Recent advances in generative modeling have introduced a new paradigm for context-conditioned modeling, where output is synthesized directly from data distributions conditioned on input specifications, particularly in protein docking \cite{somnath2023aligned, tong2023simulation}, image generation \cite{de2021diffusion, shi2023diffusion}, language-driven planning \cite{pham2025smallplan}, and text-to-speech generation \cite{chen2023schrodinger}. Within this realm, generative-based approaches have leveraged foundational probabilistic models, such as conditional variational autoencoders \cite{kingma2013auto, sohn2015learning}, normalizing flows \cite{rezende2015variational, lipman2022flow}, and diffusion \cite{ho2020denoising, song2020score}, to learn mappings from inputs (\textit{e.g.}, prompts) to image distributions. By conditioning image synthesis on contextual inputs, these models can flexibly generate new samples and produce diverse outputs that satisfy specified semantic or structural constraints, moving beyond the limitations of traditional optimization-based techniques \cite{dhariwal2021diffusion, rombach2022high}.

\begin{figure}[t]
    \centering
    \vspace{6pt}
    \includegraphics[width=1.00\linewidth]{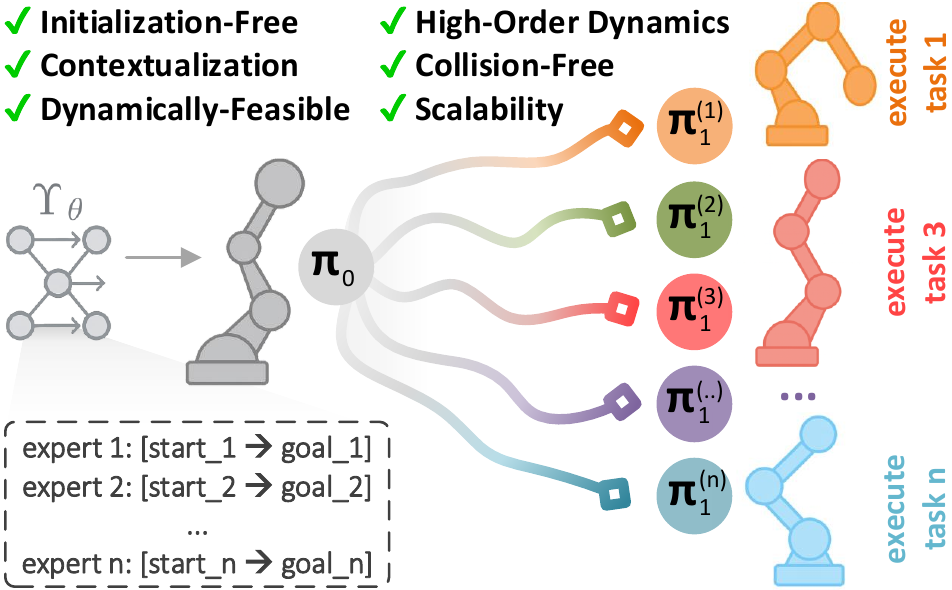}
    \vspace{-10pt}
    \caption{\textbf{Overview with $\mathbf{X}$\textit{FlowMP}:} The task-conditioned motion planner, $\Upsilon_{\theta}$, learns from diverse experts to generate trajectories based on task inputs. From an initial noise distribution $\pi_{0}$, the learned model with Schr{\"o}dinger bridges is able to produce multiple valid solutions $\pi_{1}^{(i)}$, while being initialization-free, contextualized, dynamically-feasible, collision-free, and scalable. Thus, the generated trajectories can be executed on robots across a variety of tasks.}
    \label{fig:xflowmp_modes}
    \vspace{-18pt}
\end{figure}

Extending these conditional generative models to motion planning requires producing temporally coherent trajectories that satisfy dynamic, physical, and task-specific constraints \cite{janner2022planning, chi2023diffusion}. Leveraging task contexts as semantic inputs, such as start and goal states, enables the generation of trajectories that are not only feasible and collision-free but also adaptive and diverse across various scenarios. As trajectories are high-dimensional and continuous, it also requires such models to capture long-range temporal dependencies while maintaining smoothness and feasibility \cite{tong2020trajectorynet, rubanova2019latent}. In addition, diverse tasks and environments demand robust generalization, often from limited demonstrations \cite{lynch2020learning, jang2022bc}. Moreover, the problem of complying with physical constraints such as kinematics, dynamics, and collision avoidance is typically challenging to encode in generative motion planners. 

In general, the central problem lies in modeling trajectories that are simultaneously high-dimensional, high-order dynamics, and sequential in nature, which makes capturing long-range dependencies while ensuring smoothness and feasibility particularly challenging. These difficulties are compounded by the need to integrate physical and environmental constraints, such as collision avoidance and kinematic limits. While recent advances in conditional trajectory generation have made progress \cite{janner2022planning, nguyen2025flowmp}, scalable and reliable solutions that address these intertwined challenges in a unified generative framework remain an open research problem. 

To address these limitations, we propose Schr{\"o}dinger Flow Motion Planning, namely $\mathrm{X}$\textit{FlowMP}, a task-conditioned generative framework (Fig. \ref{fig:xflowmp_modes}) that models high-dimensional robot motions as entropic flows evolving from noises to expert demonstrations. In specific, $\mathrm{X}$\textit{FlowMP} leverages motion fields with learnable score-based functions to generate smooth, dynamically-feasible, and collision-free trajectories with high-order dynamics while respecting start-goal configurations and dynamic constraints. By unifying positional, velocity, and acceleration fields with Schr{\"o}dinger bridges, our method implicitly encodes high-order dynamics and preserves the scalability of trajectories. Meanwhile, the score function guides the conditional generation in the latent space, guaranteeing a tight distributional alignment with expert demonstrations. Our contributions are stated as follows: 
\begin{enumerate}
    \item We devise a initialization-free, task-conditioned generative motion planner using Schr{\"o}dinger bridges.
    \item We unify task-conditioned motion fields with velocity, acceleration, and jerk fields for the generation of scalable, dynamically-feasible, and collision-free motions.
    \item We validate the performance of our proposed method on \texttt{RobotPointMass} and \texttt{LASA Handwriting} datasets for comprehensive evaluations and through real-world experiments on a real-robot system.
\end{enumerate}

%% file: 02_related_work.tex
\section{Related Work}

\textbf{Optimal Transport for Motion Field Synthesis:} 
Optimal transport (OT) has been shown as a strong framework for generative modeling, with most prior works applying OT, Schr{\"o}dinger bridges \cite{vargas2023denoising} and diffusion models \cite{lipman2022flow, ho2020denoising, wen2025edm} to the field of image generation \cite{de2021diffusion, shi2023diffusion} and cell synthesis \cite{somnath2023aligned, tong2023simulation}. Recent advances have extended OT-based generative models to video synthesis \cite{xu2020cot} and 3D shape generation \cite{shen2021accurate}, demonstrating the versatility of OT in high-dimensional domains. Building on these foundations, OT-based approaches have begun to address motion planning and trajectory generation in cellular synthesis, including trajectory interpolation \cite{tong2020trajectorynet, huguet2022manifold}. In robotics, conditional generative models \cite{janner2022planning, brehmer2023edgi} and neural ordinary differential equations (ODEs) \cite{rubanova2019latent, kidger2020neural} further enhance the flexibility and robustness of motion synthesis, enabling context-aware trajectory generation. To bridge the gap between image generation and trajectory synthesis, our approach seeks the derivation task-conditioned motion fields for generative motion robotic planning by implicitly contextualizing tasks and applying Schr{\"o}dinger bridge-based flow matching.

\textbf{Task-Conditioned Generative Planning:} 
The field of generative modeling has witnessed a paradigm shift driven by advances in conditional synthesis \cite{ho2022classifier, rombach2022high}. Prior works in text-to-image and text-to-video \cite{ saharia2022photorealistic, zhang2025show} have demonstrated the remarkable ability to generate high-fidelity, diverse outputs conditioned on rich semantic prompts, enabling users to specify complex visual and temporal content through natural language \cite{singer2022make}. While these generative models offer powerful semantic conditioning, directly applying them to robotics presents significant challenges due to the need for precise control, safety, and real-time adaptation in physical environments \cite{jiang2022vima, huang2025roboground}. Bridging the gap requires integrating domain knowledge, robust perception, and control mechanisms to ensure generated behaviors are feasible, reliable, and interpretable for autonomous robotic systems \cite{lynch2020learning, zeng2021transporter, jang2022bc}. Inspired by recent advances, we propose a task-conditioned generative framework for robotic motion planning. By adapting task-driven synthesis and Schr{\"o}dinger bridge-based flow matching, our method enables robots to generate scalable, collision-free, physically-feasible trajectories with high-order dynamics, which are constrained by the environment.

\textbf{Robot Planning with Schr{\"o}dinger Bridges:} 
Schr{\"o}dinger bridge theory has gained traction as a powerful tool for optimal control and distribution steering in complex robotic systems. Sahoo \textit{et al.} \cite{sahoo2018optimal} extend these ideas to the setting of Lie groups, analyzing the behavior of controlled dynamics and providing a geometric perspective on Schr{\"o}dinger-based control. Later, Bakshi \textit{et al.} \cite{bakshi2020schrodinger} introduce a Schr{\"o}dinger bridge approach for controlling large populations, demonstrating how stochastic optimal transport can be leveraged for population-level control tasks. Caluya and Halder \cite{caluya2021reflected} develop the reflected Schr{\"o}dinger bridge framework, enabling density control under explicit path constraints, which is particularly relevant for robotics and autonomous systems operating in constrained environments. Eldesoukey and Georgiou \cite{eldesoukey2024schrodinger} further generalize Schr{\"o}dinger bridge control and estimation to spatio-temporal distributions on graphs, opening new avenues for distributed and networked control. Mirrahimi \textit{et al.} \cite{mirrahimi2005lyapunov} explore Lyapunov-based control of bilinear Schr{\"o}dinger equations, providing theoretical guarantees for stability and convergence in quantum and classical systems. To push these theoretical solutions for robot planning more scalable and in a practical manner, our approach looks into the coupling between neural ODEs and score functions as learnable components for robotic planning, while preserving their high-order dynamics and scalability.

\begin{figure*}[t]    
    \centering
    \vspace{3pt}
    \includegraphics[width=1.00\linewidth]{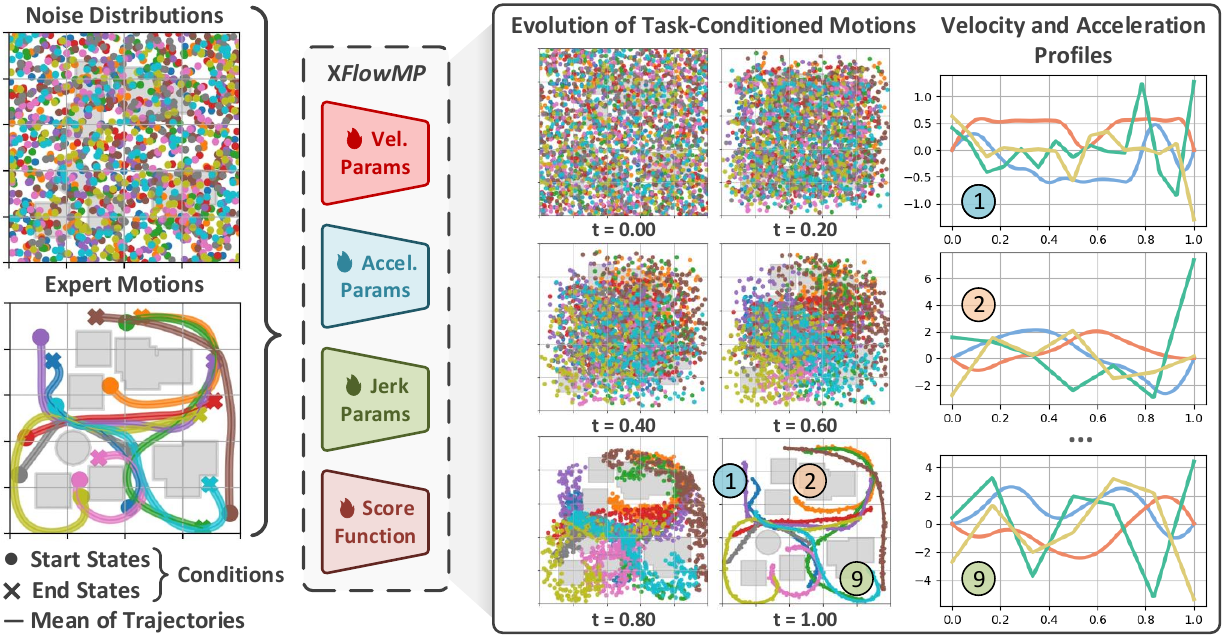}
    \vspace{-12pt}
    \caption{\textbf{Methodology of $\mathbf{X}$\textit{FlowMP}:} Given expert demonstrations of motions in the \textcolor{Gray}{\textbf{maze environment}}, $\mathrm{X}$\textit{FlowMP} contextualizes those motions by their start-goal pairs (\start,~\goal) as tasks. During training, the learnable parameters transport initial noise distributions to match expert trajectories through Schr{\"o}dinger bridges, ensuring that the generated trajectories remain consistent with task inputs. In specific, the motion field, comprising \textcolor{BrickRed}{\textbf{velocity}}, \textcolor{CornflowerBlue}{\textbf{acceleration}}, and \textcolor{LimeGreen}{\textbf{jerk}} parameters, learns to generate high-order motions along the evolution horizon from $0$ to $1$. Meanwhile, the \textcolor{Maroon}{\textbf{score function}} estimates the underlying task-dependent gradients that guide the flows toward expert distributions. The corresponding velocity (\textcolor{lightblue}{\textbf{blue}} and \textcolor{lightorange}{\textbf{orange}}) and acceleration (\textcolor{tealgreen}{\textbf{green}} and \textcolor{lightyellow}{\textbf{yellow}}) profiles of the paths are shown accordingly. At $t = 1.00$, $\mathrm{X}$\textit{FlowMP} generates motions that are collision-free and dynamically-feasible while maintaining semantic consistency with task conditions.}
    \vspace{-18pt}
    \label{fig:xflowmp_evolution}
\end{figure*}

%% file: 03_methodology.tex
\section{Schr{\"o}dinger Flow Motion Planning}
\label{sec:methodology}
\subsection{Problem Formulation}
Consider a workspace $\mathcal{W}$ in which a robot can navigate from a start configuration to a goal configuration while avoiding obstacles and satisfying dynamic constraints. We define the set of $N$ expert demonstration motions as:
\begin{equation*}
    \mathcal{D} = \left\{ \pi^{(i)} = \left\{\mathbf{q}^{(i)}_{t}, \mathbf{\dot{q}}^{(i)}_{t}, \mathbf{\ddot{q}}^{(i)}_{t} \right\}_{t=0}^{T} \; \middle| \; \mathbf{q}^{(i)}_{0} \in \mathcal{S}, \; \mathbf{q}^{(i)}_{T} \in \mathcal{G} \right\},
    \label{eq:expert_demos}
\end{equation*}
where $\pi^{(i)}$ denotes the motion for the $i$-th task, $\mathbf{q}^{(i)}_{t} \in \mathbb{R}^d$ is the robot's state at time step $t$ along the planning horizon from $0$ to $T$, $\mathcal{S} \subset \mathcal{W}$ and $\mathcal{G} \subset \mathcal{W}$ denote the sets of start and goal states within the robot workspace, respectively.

To learn the generative model for robotic motion planning, we first define the distributions of expert trajectories as positional states, $\mathbf{Q}$, and their velocity and acceleration states, $\mathbf{\dot{Q}}$ and $\mathbf{\ddot{Q}}$, respectively. We denote $\pi_{0} = \{ \varepsilon_{\mathbf{Q}}, \varepsilon_{\mathbf{\dot{Q}}}, \varepsilon_{\mathbf{\ddot{Q}}} \}$ and $\pi_{1} = \{ \mathbf{Q}, \mathbf{\dot{Q}}, \mathbf{\ddot{Q}} \}$ as the noise and target motion distributions, respectively, and utilize flows to transport from the source distribution $\pi_{0}$ to the target distribution $\pi_{1}$. Given such an expert dataset $\mathcal{D}$, our goal thus is to learn a generative planner that can synthesize scalable, collision-free, and dynamically feasible trajectories with high-order dynamics under multiple start-goal configurations within the robot workspace. The neural planner can be formally written as follows:
\begin{equation}
    \Upsilon_{\theta} \left(\mathbf{q}_{0}, \mathbf{q}_{T}, \pi_{0} \mid \mathcal{D}, \mathcal{W} \right) \rightarrow \pi'_{1} \text{, with } \mathbf{q}_{0} \in \mathcal{S}, \; \mathbf{q}_{T} \in \mathcal{G} 
    \label{eq:problem_formulation}
\end{equation}
where $\Upsilon_{\theta}$ represents the learnable motion field with learnable parameters $\theta$ and $\pi'_{1} = \{\mathbf{q}'_{t}\}_{t=0}^{T} \notin \mathcal{D}$ denotes the generated trajectory that is conditioned by the start-goal states $\left( \mathbf{q}_{0}, \mathbf{q}_{T} \right)$ within the robot workspace $\mathcal{W}$. From here, we refer to the start-goal states as a task, denoted as $\mathcal{T}$, for simplicity.

\subsection{Motion Field with Schr{\"o}dinger Bridges}
Given a reference stochastic process $\mathbb{Q}$ with law on motions of $\pi$, the Schr{\"o}dinger bridge finds the motion distribution $\mathbb{P}^{\star}$ closest to $\mathbb{Q}$ via Kullback–Leibler (KL) divergence, matching the source distribution $\pi_{0}$ to target distribution $\pi_{1}$:
\begin{equation}
    \mathbb{P}^{*} = \arg \min_{\mathbb{P}:\; \mathbb{P}_{0} = \pi_{0}, \mathbb{P}_{1} = \pi_{1}} \; \mathrm{KL} \left( p_{t}(\pi) \mid\mid q_{t} \right),
    \label{eq:schrodinger_brige}
\end{equation}
where $p_t(\pi) \in \mathbb{P}$ represents the evolving probability flows and $q_{t} \in \mathbb{Q}$ is a reference measure of the stochastic process. 

From the vanilla motion field $\Upsilon$ of the time-dependent flow $\psi_{t}$ \cite{nguyen2025flowmp}, the stochastic differential equation (SDE) of a motion field under entropic flows $\Upsilon \left( \psi(\pi) \right)$ is defined as:
\begin{equation}
    d\psi_{t} = \Upsilon_{t}\left(\psi(\pi) \right) dt + \sigma_{t} dw_{t},
    \label{eq:motion_field_sde}
\end{equation}
where $\psi$ denotes the entropic flows and $w$ models a time-independent Brownian motion with diffusion strength of $\sigma$. Also, based on the well-known Fokker-Planck equation \cite{fokker1914mittlere, planck1917satz}, the continuity equation is given as:
\begin{equation}
    \frac{dp(\pi)}{dt} = -\nabla \cdot \left[ p_t(\pi) \Upsilon_t \left(\psi(\pi) \right) \right] + \tfrac{1}{2} \sigma_t^{2} \Delta p_{t}(\pi),
    \label{eq:kolmogorov}
\end{equation}
where the advection term describes the transport of probability mass and the diffusion term models the spreading effect of noise. Thus, to obtain an ODE that preserves the same marginal distributions $p_{t}(\pi)$ in Eq. \ref{eq:kolmogorov}, we rewrite the probability flow ODE from its SDE in Eq. \ref{eq:motion_field_sde}:
\begin{equation}
    \Upsilon'_{t} \left(\psi(\pi)\right) 
    = \Upsilon_{t} \left(\psi(\pi)\right) - \tfrac{1}{2} \sigma_{t}^{2} \nabla \log p_{t}(\pi),
    \label{eq:new_drift}
\end{equation}
where $\Upsilon^{'}_{t} \left( \psi(\pi) \right)$ is the equivalent drift function that ensures that $p_t(\pi)$ evolves as in Eq. \ref{eq:kolmogorov}, and $\nabla \log p_t(\pi)$ is the score function. Under the linear-Gaussian reference processes in flow-matching settings, the marginals, $p_t(\pi) = \mathcal{N}\left(\mu_{\pi}, \sigma^{2}_{\pi}\right)$, admit the following closed form with the Schr{\"o}dinger bridges from $\pi_0$ to $\pi_1$ and a constant diffusion rate $\sigma \equiv \sigma_{t}$:
\begin{equation}
    p_{t}(\pi) = \mathcal{N} \left( t\pi_{1} + (1-t)\pi_{0}, \; \sigma^{2} t (1-t) \right).
    \label{eq:flow_marginals}
\end{equation}
From Eq. \ref{eq:new_drift} and Eq. \ref{eq:flow_marginals}, the motion field ODE and the score function are thus expressed as:
\begin{subequations}
    \begin{equation}
        \Upsilon'_{t} \left( \psi(\pi) \right) = \alpha(t) (\pi - \mu_{\pi}) + (\pi_{1} - \pi_{0}),
        \label{eq:motion_field_ode}
        \vspace{-4pt}
    \end{equation}
    \begin{equation}
        \nabla \log p_{t}(\pi) = (\mu_{\pi} -\pi)/\sigma^{2}_{\pi},
        \label{eq:score_function}
    \end{equation}
\end{subequations}
where $\alpha(t)$ is a transport weighting term induced by the interpolants. Here, Eq. \ref{eq:motion_field_ode} and Eq. \ref{eq:score_function} provide the Schr{\"o}dinger bridge-based motion field in unconditioned settings.

\setlength{\textfloatsep}{4pt}
\begin{algorithm}[t]
    \caption{Schr{\"o}dinger Flow Motion Planning}
    \label{alg:schrodinger_flow_motion_planning}
    \begin{normalsize}
    \DontPrintSemicolon
    \SetKwInOut{KwIn}{Input}
    \SetKwInOut{KwOut}{Output}
    \SetKwFunction{FTrain}{\fontfamily{lmtt}\selectfont train\_xflowmp}
    \SetKwFunction{FInfer}{\fontfamily{lmtt}\selectfont generate\_motion}
    \SetKwProg{Fn}{function}{}{}
    \KwIn{$\pi_{1} = \{\mathbf{Q}, \mathbf{\dot{Q}}, \mathbf{\ddot{Q}}\} \coloneqq$ expert motions, \\
          $\pi_{0}= \{\varepsilon_{\mathbf{Q}}, \varepsilon_{\mathbf{\dot{Q}}}, \varepsilon_{\mathbf{\ddot{Q}}}\} \coloneqq$ noise distributions, \\
          $\sigma \coloneqq$ diffusion strength, $\lambda(t) \coloneqq$ scheduler}
    \KwOut{$\Upsilon_{\theta} \coloneqq$ motion field, $g_{\theta} \coloneqq$ score function}

    \Fn{\FTrain{$\pi_{1}, \pi_{0}, \sigma, \lambda(t)$}}{
        \While{training}{
            Sample $t \sim \mathcal{U}[0,1]$ \\
            Sample expert minibatch $(\mathbf{q}_{1},\mathbf{\dot{q}}_{1},\mathbf{\ddot{q}}_{1}) \sim \pi_{1}$ \\
            Sample noise minibatch $(\varepsilon_{\mathbf{q}}, \varepsilon_{\mathbf{\dot{q}}}, \varepsilon_{\mathbf{\ddot{q}}}) \sim \pi_{0}$ \\
            $\mathcal{T} \gets \texttt{\small get\_start\_goal\_states}(\mathbf{q}_{1})$ \\
            $\mathbf{q}_{t} \gets \texttt{\small interp}(\mathbf{q}_{1}, \varepsilon_{\mathbf{q}}, \varepsilon_{\mathbf{\dot{q}}}, \varepsilon_{\mathbf{\ddot{q}}}, t)$ \text{ (Eq. \ref{eq:interpolants})} \\
            $\Upsilon_{t} \gets \texttt{\small get\_fields}(\mathbf{q}_{1}, \mathbf{\dot{q}}_{1}, \mathbf{\ddot{q}}_{1}, \varepsilon_{\mathbf{q}}, \varepsilon_{\mathbf{\dot{q}}}, \varepsilon_{\mathbf{\ddot{q}}}, t)$ \\
            \textcolor{gray}{// compute motion field drift and task score} \\
            $\Upsilon'_{t} \gets \Upsilon'_{t}(\mathbf{q}_{t} \mid \mathcal{T})$ \text{ (Eq. \ref{eq:task_dependent_motion_field_ode})} \\
            $s_{t} \gets \nabla \log p_{t}(\mathbf{q}_{t} \mid \mathcal{T})$ \text{ (Eq. \ref{eq:task_dependent_score_function})} \\ 
            \textcolor{gray}{// train neural motion field and score function} \\
            $\widehat{\Upsilon}_{t} \gets \Upsilon_{\theta}(\mathbf{q}_{t}, t)$, \quad $\widehat{s}_{t} \gets s_{\theta}(\mathbf{q}_{t}, t)$ \\ 
            $\mathcal{L}(\theta) \gets \|\widehat{\Upsilon}_{t} - \Upsilon'_t\|_{2}^{2} + \lambda(t)\,\|\widehat{s}_{t} - s_t\|_{2}^{2}$ \text{(Eq. \ref{eq:total_loss})} \\
            $\theta \leftarrow \theta - \eta \nabla_{\theta}\mathcal{L}(\theta)$ \\
        }
        \KwRet{$\Upsilon_{\theta}, s_{\theta}$}
    }

    \vspace{4pt}
    \KwIn{$\pi_{0} \coloneqq$ noise distributions, $\delta \coloneqq$ timestep \\ 
    $\mathcal{T} = [\mathbf{q}_{0}, \mathbf{q}_{T}] \coloneqq$ task (start \& goal states)}
    \KwOut{$[\mathbf{q}, \mathbf{\dot{q}}, \mathbf{\ddot{q}}] \coloneqq$ task-conditioned robot motion}
    \Fn{\FInfer{$\pi_{0}, \delta, \mathcal{T}$}}{
        $\mathbf{q}, \mathbf{\dot{q}}, \mathbf{\ddot{q}} = \varepsilon_{\mathbf{Q}}, \varepsilon_{\mathbf{\dot{Q}}}, \varepsilon_{\mathbf{\ddot{Q}}} \gets \pi_{0}$, \quad $t \gets 0$ \\
        \While{$t \leq 1$}{
            $s \gets s_{\theta}(t, \mathbf{q}, \mathbf{\dot{q}}, \mathbf{\ddot{q}}, \mathcal{T})$ \\
            $\Upsilon \gets \Upsilon_{\theta}(t, \mathbf{q}, \mathbf{\dot{q}}, \mathbf{\ddot{q}}, \mathcal{T}) - \tfrac{1}{2}\,\sigma_{t}^{2} s$ \text{ (Eq. \ref{eq:motion_field_update})} \\
            $[\mathbf{q}, \mathbf{\dot{q}}, \mathbf{\ddot{q}}] \gets [\mathbf{q}, \mathbf{\dot{q}}, \mathbf{\ddot{q}}] + \Upsilon \cdot \delta$ \\
            $t \gets t + \delta$ \\
        }
        \KwRet{$[\mathbf{q},\mathbf{\dot{q}},\mathbf{\ddot{q}}]$}
    }
    \end{normalsize}
\end{algorithm}

\subsection{Task-Conditioned Motion Field with Schr{\"o}dinger Bridges}
Two important aspects of realizing task-conditioned robot motions are score-based functions and motion fields. The score-based function serves as a guided sampling mechanism post-contextualization. Meanwhile, the motion field refers to the generation of a trajectory given start and goal states.

Given the set of $N$ distinct tasks $\mathcal{T} = \bigcup_{i=1}^{N} \mathcal{T}_{i}$, Eq. \ref{eq:flow_marginals} with any task $\mathcal{T}$ can be modeled as:
\begin{equation}
     p_{t} \left( \pi \mid \mathcal{T} \right) = \mathcal{N} \left( t\pi^{\mathcal{T}}_{1} + (1-t) \pi^{\mathcal{T}}_{0}, \; \sigma_{\mathcal{T}}^{2} t(1-t) \right),
    \label{eq:task_dependent_flow_marginals}
\end{equation}
with $\bigcup_{i=1}^{N} \pi^{\mathcal{T}_{i}} \subseteq \pi$, $\bigcup_{i=1}^{N} \pi^{\mathcal{T}_i}_{0} \subseteq \pi_{0}$, and $\bigcup_{i=1}^{N} \pi^{\mathcal{T}_i}_{1} \subseteq \pi_{1}$ representing the task-dependent intermediate, source, and expert motion distributions along the evolving probability flow $p_{t}(\pi \mid \mathcal{T})$, respectively. With Eq. \ref{eq:task_dependent_flow_marginals}, the motion field ODE (Eq. \ref{eq:motion_field_ode}) and score function (Eq. \ref{eq:score_function}) are then adjusted to reflect the task dependency as follows:
\begin{subequations}
    \begin{equation}
        \Upsilon'_{t} \left( \psi \left( \pi \right) \mid \mathcal{T} \right) = \alpha(t) \left( \pi - \mu_{\pi^{\mathcal{T}}} \right) + \left( \pi^{\mathcal{T}}_{1} - \pi^{\mathcal{T}}_{0} \right),
        \label{eq:task_dependent_motion_field_ode}
        \vspace{-4pt}
    \end{equation}
    \begin{equation}
        \nabla \log p_{t} \left(\pi \mid \mathcal{T} \right) = \left( \mu_{\pi^{\mathcal{T}}} - \pi^{\mathcal{T}} \right) / \sigma^{2}_{\pi^{\mathcal{T}}}.
        \label{eq:task_dependent_score_function}
    \end{equation}
\end{subequations}

Eq. \ref{eq:task_dependent_motion_field_ode} exemplifies the incorporation of task-dependent variations in transport behavior while still being grounded with Schr{\"o}dinger bridges. Meanwhile, Eq. \ref{eq:task_dependent_score_function} characterizes the score-based function that guides the trajectory generation in latent space to maximize task likelihood. To preserve high-order dynamics, we also incorporate time-dependent interpolants for path, velocity, and acceleration profiles, while training the motion field $\Upsilon_{t}$ \cite{nguyen2025flowmp}:
\begin{subequations}
    \begin{align}
        \mathbf{q}_t &= (1 - t)\,\varepsilon_{\mathbf{q}} + t\,\mathbf{q}_1, \\
        \mathbf{q}_t &= (1 - t^2)\, \varepsilon_{\mathbf{q}} + (t - t^2)\, \varepsilon_{\dot{\mathbf{q}}} + t^2\,\mathbf{q}_1, \\
        \mathbf{q}_t &= (1 - t^3)\, \varepsilon_{\mathbf{q}} + (t - t^3)\, \varepsilon_{\dot{\mathbf{q}}} 
        + \tfrac{t^2 - t^3}{2}\, \varepsilon_{\ddot{\mathbf{q}}} + t^3\, \mathbf{q}_{1},
    \end{align}
    \label{eq:interpolants}
\end{subequations}
where $\mathbf{q}_{t}$ acts as the evolving probability flow of $\mathbf{q}$, while transporting it from $\varepsilon_{\mathbf{q}} \in \varepsilon_{\mathbf{Q}}$ to $\mathbf{q}_{1} \in \mathbf{Q}$, and $\varepsilon_{\dot{\mathbf{q}}} \in \varepsilon_{\mathbf{\dot{Q}}}$ and $\varepsilon_{\ddot{\mathbf{q}}} \in \varepsilon_{\mathbf{\ddot{Q}}}$ denote the noise initializations for first- and second-order dynamics of $\mathbf{q}$ along the intermediate distribution $\pi$.

To learn both the score-based function and the motion field simultaneously, we employ two U-Nets \cite{ronneberger2015u} corresponding to Eq. \ref{eq:task_dependent_motion_field_ode} and Eq. \ref{eq:task_dependent_score_function} that jointly learn contextualized evolving entropic flows $p_{t} \left(\pi \mid \mathcal{T} \right)$ from  $\pi^{\mathcal{T}}_{0}$ to $\pi^{\mathcal{T}}_{1}$. 
We train two neural SDEs to approximate (i) the motion field $\Upsilon_{t}$ and (ii) the score function $\nabla \log p_{t}(\cdot \mid \mathcal{T})$ through the combined loss of flow-matching and score-matching, thereby solving Eq. \ref{eq:schrodinger_brige}:
\begin{equation}
    \mathcal{L}(\theta) 
    = \|\Upsilon_{\theta} - \Upsilon^{'}_{t}\|_{2}^{2} 
    + \lambda(t)\, \|s_{\theta} - \nabla \log p_t(\pi \mid \mathcal{T})\|_{2}^{2},
    \label{eq:total_loss}
\end{equation}
where $\lambda(t)$ represents the scheduler, $\Upsilon_{\theta}$ is the learned motion field, and $s_{\theta}$ is the learned score function. The motion field $\Upsilon_{\theta}$ also learns the sub-fields for high-order dynamics via the learnable parameters $\theta = \{\theta_{1}, \theta_{2}, \theta_{3}\}$, as detailed in Eq. \ref{eq:interpolants}.

\subsection{Inference of Task-Conditioned Motion Field}
After training governed by Eq. \ref{eq:total_loss}, to infer the optimal task-conditioned motion field, we estimate the transport paths by updating the transport drift and the score function. The entropic flow follows the motion field ODE: $d\psi_{t} = \Upsilon'_t \left( \psi \left( \pi \right) \mid \mathcal{T} \right) dt$, which moves the samples $\pi^{\mathcal{T}}$ from $\pi^{\mathcal{T}}_{0}$ to $\pi^{\mathcal{T}}_{1}$. With $k$ as the number of update steps in the time range from $0$ to $1$, the motion field with the score function that pushes noise samples $\pi_{0}$ toward high-likelihood expert motions while incorporating the task condition $\mathcal{T}$ is:
\begin{equation} 
    \Upsilon^{(k+1)}_{t} \left( \psi \left( \pi \right) \mid \mathcal{T} \right) = \Upsilon^{(k)}_{\theta} \left(t, \psi \left( \pi \right) \mid \mathcal{T} \right) - \tfrac{1}{2} \sigma^{2} s_{\theta} \left(t, \pi \mid \mathcal{T} \right). 
    \label{eq:motion_field_update}
\end{equation} 

\begin{table*}[t]
    \centering
    \vspace{4pt}
    \caption{Quantitative comparisons of $\mathrm{X}$\textit{FlowMP} against other flow matching-based methods \cite{lipman2022flow, tong2023improving, albergo2023stochastic} in terms of maximum mean discrepancy (MMD), trajectory jerkiness (TJ), and energy consumption (EC). For all benchmarked methods, the lengths of generated trajectories are $256$ in the \texttt{RobotPointMass} environment \cite{nguyen2025flowmp} and $1,000$ in the \texttt{LASA Handwriting} dataset \cite{lemme2015open}, respectively.}
    \vspace{-3pt}
    \begin{tabular}{r ccc ccc}
        \toprule
        \multirow{2}{*}{\diagbox{Method}{\makecell{Environment \\ \qquad / Metric}}} & \multicolumn{3}{c}{\texttt{\small RobotPointMass (L = 256)}} & \multicolumn{3}{c}{\texttt{\small LASA Handwriting (L = 1,000)}} \\
        \cmidrule(lr){2-4} \cmidrule(lr){5-7}
        & \makecell{MMD ($\downarrow$) \\(Eq. \ref{eq:metric_mmd})} & \makecell{$\mathrm{TJ}$ ($\downarrow$) \\(Eq. \ref{eq:metric_trajectory_smoothness})} & \makecell{$\mathrm{EC}$ ($\downarrow$) \\(Eq. \ref{eq:metric_energy_consumption})} & \makecell{MMD ($\downarrow$) \\(Eq. \ref{eq:metric_mmd})} & \makecell{$\mathrm{TJ}$ ($\downarrow$) \\(Eq. \ref{eq:metric_trajectory_smoothness})} & \makecell{$\mathrm{EC}$ ($\downarrow$) \\(Eq. \ref{eq:metric_energy_consumption})} \\
        \midrule
            Lipman \textit{et al.} \cite{lipman2022flow} & 12590.54 & $101.58 \pm 7.55$ & $8695.78 \pm 534.74$ & 39.88 & $2507.17 \pm 176.03$ & $215263.84 \pm 12407.40$ \\
            Tong \textit{et al.} \cite{tong2023improving} & 34.63 & $0.17 \pm 0.10$ & $15.93 \pm 8.60$ & 20.18 & $2.78 \pm 1.73$ & $231.40 \pm 158.54$ \\
            Albergo \textit{et al.} \cite{albergo2023stochastic} & 31.42 & $0.11 \pm 0.06$ & $9.83 \pm 4.65$ & 19.07 & $4.97 \pm 4.39$ & $432.17 \pm 360.72$\\
            \hline
            \rowcolor{gray!25} $\mathrm{\textbf{X}}$\textit{\textbf{FlowMP}} \textbf{(Ours)} & \textbf{14.52} & \textbf{0.07 $\pm$ 0.04} & \textbf{5.91} $\pm$ \textbf{3.63} & \textbf{18.83} & \textbf{2.67} $\pm$ \textbf{1.31} & \textbf{157.43} $\pm$ \textbf{97.16} \\
            \rowcolor{gray!15} \textit{vs.} 2nd Best & \boldgreen{+53.79\% $\uparrow$} & \boldgreen{+36.36\% $\uparrow$} & \boldgreen{+39.88\% $\uparrow$} & \boldgreen{+1.26\% $\uparrow$} & \boldgreen{+3.96\% $\uparrow$} & \boldgreen{+31.97\% $\uparrow$} \\
        \bottomrule
    \end{tabular}
    \label{tab:comparison_results}
    \vspace{-14pt}
\end{table*}

By applying $k$-step optimization, Eq. \ref{eq:motion_field_update} is able to generate smooth, scalable, collision-free, dynamically-feasible, task-adaptive robot motions with high-order dynamics under contextualization. We outline our proposed method, as illustrated in Fig. \ref{fig:xflowmp_evolution} and described algorithmically in Alg. \ref{alg:schrodinger_flow_motion_planning}.



%% file: 04_evaluation.tex
\section{Evaluations \& Experiments}
\label{sec:evaluations}
\subsection{Evaluation Baselines \& Questions}
\subsubsection{Datasets \& Baselines}
To assess $\mathrm{X}$\textit{FlowMP}'s performance comprehensively, we evaluate our proposed method in terms of distributional discrepancy, trajectory jerkiness, and curves with least energy, on the \texttt{RobotPointMass} dataset \cite{nguyen2025flowmp}, where expert motions are generated using VP-STO \cite{jankowski2022vp}, and the \texttt{LASA Handwriting} dataset \cite{lemme2015open} with high-fidelity written character writings. The experimented baselines include three flow matching variants:
\begin{itemize}
    \item Lipman \textit{et al.} \cite{lipman2022flow}: A foundational work that aligns generated samples directly to targets through flows.
    \item Tong \textit{et al.} \cite{tong2023improving}: A variant that leverages the theory of optimal transport to compute exact displacement fields between source and target distributions.
    \item Albergo \textit{et al.} \cite{albergo2023stochastic}: Another approach that introduces stochastic flows with trigonometric interpolants.
\end{itemize}

\subsubsection{Evaluation Questions}
All algorithms are benchmarked using the same network architecture, trained for $30$ epochs on the NVIDIA RTX 4090 GPU with initialized noises randomly generated within the robot workspace. Our evaluations are to answer the following key questions:
\begin{itemize}
    \item \textbf{Q1:} How well do generated trajectories by $\mathrm{X}$\textit{FlowMP} match the expert trajectory distribution?
    \item \textbf{Q2:} Are those generated trajectories smoother or jerkier compared to those generated by baseline methods?
    \item \textbf{Q3:} Given the first two criteria, whether the generated trajectories by our proposed method follow paths of lower curve-energy compared to baseline methods?
    \item \textbf{Q4:} Does Schr{\"o}dinger-based approach yield better contextualization and scalability while generating collision-free and dynamically-feasible motions?
    \item \textbf{Q5:} Across the benchmarked algorithms, how do the generated motions compare to each other qualitatively?
\end{itemize}

\subsection{Quantitative Results} 
\subsubsection{Maximum Mean Discrepancy (MMD)}
First, to measure the distance between the generated and the learning distributions in order to answer \textbf{Q1}, we use MMD \cite{gretton2012kernel} to evaluate their discrepancy in a reproducing kernel Hilbert space. Let $\mathcal{D}_{\text{gen}} = \{\mathbf{q}_{i}^{\text{gen}}\}_{i=1}^{M}$ be the set of $M$ generated trajectories, the MMD score is defined as:
\begin{equation*}
    \text{MMD} \left( \mathcal{D}_{\text{gen}}, \mathcal{D} \right) = \frac{1}{M^{2}} \sum_{i, i'} k \left( \mathbf{q}_{i}^{\text{gen}}, \mathbf{q}_{i'}^{\text{gen}} \right)
\end{equation*}
\vspace{-7pt}
\begin{equation}
    - \frac{2}{MN} \sum_{i,j} k( \left( \mathbf{q}_{i}^{\text{gen}}, \mathbf{q}_{j} \right) + \frac{1}{N^2} \sum_{j,j'} k \left( \mathbf{q}_{j}, \mathbf{q}_{j'} \right),
    \label{eq:metric_mmd}
\end{equation}
where $k(\cdot, \cdot)$ represents the positive definite kernel.

In the \texttt{RobotPointMass} environment, $\mathrm{X}$\textit{FlowMP} achieves the lowest MMD of $14.52$, outperforming prior flow matching variants, including Albergo \textit{et al.} and Tong \textit{et al.}, as shown in Tab. \ref{tab:comparison_results}. Meanwhile, on the more challenging \texttt{LASA Handwriting} dataset, $\mathrm{X}$\textit{FlowMP} retains a lower MMD of $18.83$, slightly below Albergo \textit{et al.} and surpassing Tong \textit{et al.}. Overall, these MMD scores demonstrate that $\mathrm{X}$\textit{FlowMP} is able to maintain distributional consistency with respect to expert motions for both short- and long-horizon planning.

\begin{figure*}[t]    
    \centering
    \vspace{4pt}
    \includegraphics[width=1.00\linewidth]{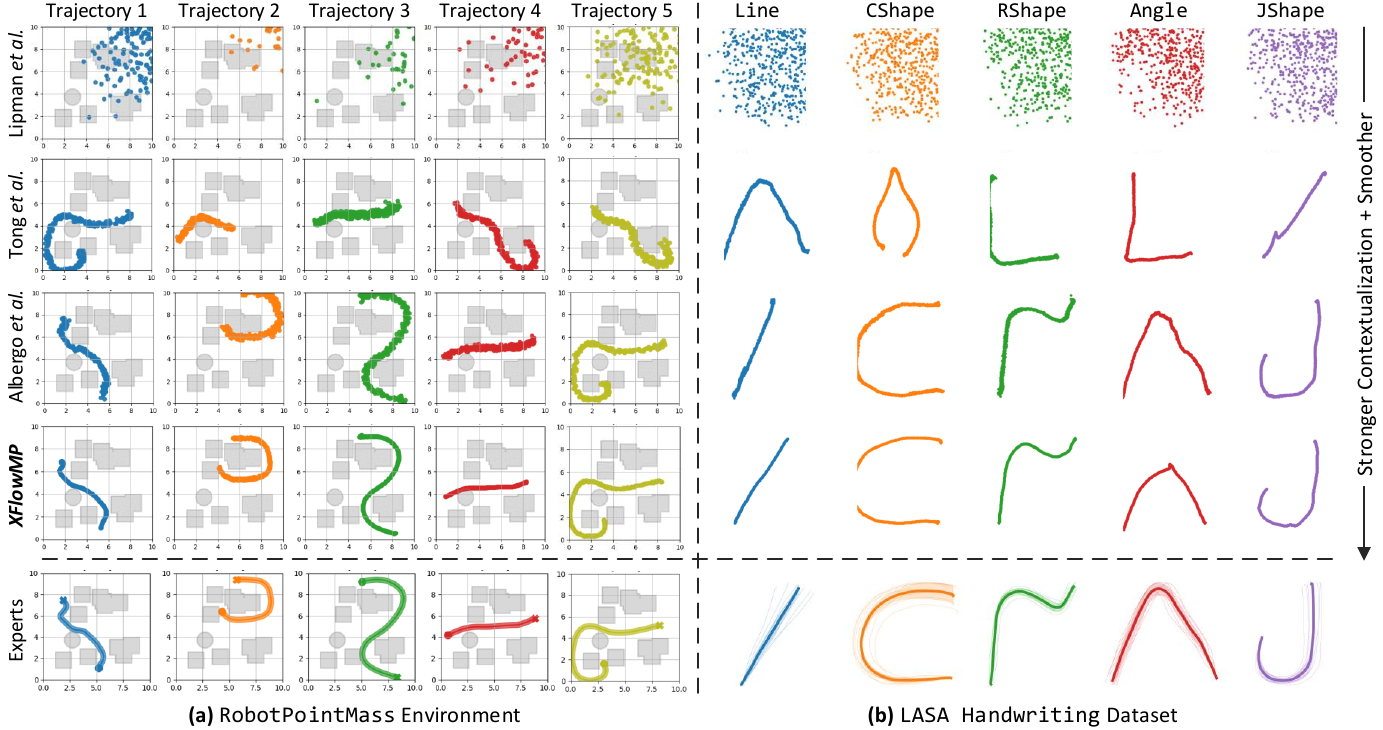}
    \vspace{-14pt}
    \caption{\textbf{Qualitative Comparisons of Trajectories Generated by Baselines and $\mathbf{X}$\textit{FlowMP}:} \textbf{(a)} In the \texttt{RobotPointMass} environment, $\mathrm{X}$\textit{FlowMP} produces smooth, collision-free paths that closely follow expert demonstrations, while baselines often yield discontinuous or collision-prone trajectories. \textbf{(b)} On the \texttt{LASA Handwriting} dataset, $\mathrm{X}$\textit{FlowMP} captures the structure of complex patterns, aligning well with expert motions. In contrast, baseline methods tend to generate incorrect/distorted shapes or fail to preserve curvature. Generally, $\mathrm{X}$\textit{FlowMP} exhibits a strong ability to contextualize while producing smooth, collision-free, and expert-like trajectories.} 
    \vspace{-18pt}
    \label{fig:xflowmp_quality}
\end{figure*}

\subsubsection{Trajectory Jerkiness}
Next, to quantify the jerkiness of the generated trajectory for \textbf{Q2}, we compute the integrated squared acceleration along the trajectory. Along the planning horizon $T$, we calculate the trajectory jerkiness as:
\begin{equation}
    \text{TJ} \left(\mathbf{q} \right) = \int_{0}^{T} \|\ddot{\mathbf{q}}(t)\|_{2}^{2} dt,
    \label{eq:metric_trajectory_smoothness}
\end{equation}

Tab. \ref{tab:comparison_results} also shows that $\mathrm{X}$\textit{FlowMP} consistently produces the lowest values on both datasets. In the \texttt{RobotPointMass} environment, our method attains $0.07 \pm 0.04$, a substantial improvement over Albergo \textit{et al.} and Tong \textit{et al.}. On the \texttt{LASA Handwriting} dataset, $\mathrm{X}$\textit{FlowMP} again achieves the best performance with $2.67 \pm 1.31$, improving upon Tong \textit{et al.} ($2.78 \pm 1.73$) and Albergo \textit{et al.} ($4.97 \pm 4.39$). Through these jerkiness scores, we confirm that our method generates smoother and physically-feasible motions.

\subsubsection{Energy Consumption}
To answer \textbf{Q3}, we assess the efficiency of the generated trajectory by measuring its total kinetic energy along the horizon $T$, as follows:
\begin{equation}
    \text{EC} \left( \mathbf{q} \right) = \int_{0}^{T} \|\dot{\mathbf{q}}(t)\|_{2}^{2} dt, 
    \label{eq:metric_energy_consumption}
\end{equation}

Given the correctness of the first two criteria, $\mathrm{X}$\textit{FlowMP} also achieves the lowest energy consumption among the benchmarked methods, as depicted in Tab. \ref{tab:comparison_results}. In the \texttt{RobotPointMass} environment, our method retains the energy consumption of $5.91 \pm 3.63$, which is substantially lower than Albergo \textit{et al.} and Tong \textit{et al.}. Similarly, on \texttt{LASA Handwriting} dataset, $\mathrm{X}$\textit{FlowMP} produces the most energy-efficient trajectories with $157.43 \pm 97.16$, much lower than Tong \textit{et al.} and Albergo \textit{et al.}, further indicating that $\mathrm{X}$\textit{FlowMP} generates trajectories that are not only smooth but also energy efficient compared to other baselines.

\subsubsection{Summary}
Taken together, these results showcase that $\mathrm{X}$\textit{FlowMP} yields trajectories that achieve superior distributional alignment, smoothness, and energy efficiency compared to state-of-the-art flow matching baselines. Notably, our method demonstrates effective scalability across a range of horizons, from short to long, showing strong adaptability to the requirements of trajectory resolutions.

\begin{table}[t]
    \centering
    \vspace{3pt}
    \caption{Planning feasibility for both short- and long-horizon motions among all flow matching-based motion planning baselines. \cmark~denotes success of and \xmark~denotes failure to reach targets. Also, the number of successful task contextualizations is reported.}
    \vspace{-2pt}
    \begin{tabular}{r cc}
        \toprule
        \diagbox{Method}{Length} & \makecell{\texttt{RobotPointMass} \\ \texttt{(L = 256)}} & \makecell{\texttt{LASA Handwriting} \\ \texttt{(L = 1,000)}} \\
        \midrule \midrule
        Lipman \textit{et al.} \cite{lipman2022flow} & 0/10 (\xmark) & 0/10 (\xmark) \\
        Tong \textit{et al.} \cite{tong2023improving} & 5/10 (\xmark) & 7/10 (\xmark) \\
        Albergo \textit{et al.} \cite{albergo2023stochastic} & \textbf{10/10} (\cmark) & \textbf{10/10} (\cmark) \\
        \hline
        \rowcolor{gray!25} $\mathrm{\textbf{X}}$\textit{\textbf{FlowMP}} \textbf{(Ours)} & \textbf{10/10} (\cmark) & \textbf{10/10} (\cmark) \\
        \bottomrule
    \end{tabular}
    \label{tab:planning_feasibility}
\end{table}

\subsection{Planning Feasibility \& Inference time}
\subsubsection{Planning Feasibility}
To quantitatively answer \textbf{Q4}, we evaluate planning feasibility based on the number of successful planned paths over the tested paths, as shown in Tab. \ref{tab:planning_feasibility}. The results of feasibility show that both $\mathrm{X}$\textit{FlowMP} and Albergo \textit{et al.} reach perfect success rates, producing collision-free and goal-reaching trajectories in both short- and long-horizon tasks. Lipman \textit{et al.} fails in every test case, while Tong \textit{et al.} only manages partial success, with $5/10$ valid trajectories in short-horizon and $7/10$ in long-horizon planning. $\mathrm{X}$\textit{FlowMP} demonstrates strong reliability across different planning horizons and maintains consistent performance compared to the state-of-the-art baselines, which also yields that the usage of Schr{\"o}dinger bridges is more suitable for task-adaptive planning conditions and requirements compared to other methods.

\begin{table}[t]
    \centering
    \vspace{4pt}
    \caption{Benchmarking the planning/inference time (in \textit{seconds}) for both short- and long-horizon motions among baselines.}
    \vspace{-3pt}
    \begin{tabular}{r cc}
        \toprule
        \diagbox{Method}{Time} & \makecell{Planning Time \\ \texttt{(L = 256)}} & \makecell{Planning Time \\ \texttt{(L = 1,000)}} \\
        \midrule \midrule
        Lipman \textit{et al.} \cite{lipman2022flow} & -- & -- \\
        Tong \textit{et al.} \cite{tong2023improving} & 0.1911 $\pm$ 0.0240 & \textbf{0.1973 $\pm$ 0.0364} \\
        Albergo \textit{et al.} \cite{albergo2023stochastic} & 0.2477 $\pm$ 0.0310 & 0.2859 $\pm$ 0.0817 \\
        \hline
        \rowcolor{gray!25} $\mathrm{\textbf{X}}$\textit{\textbf{FlowMP}} \textbf{(Ours)} & \textbf{0.1687 $\pm$ 0.0393} & 0.2577 $\pm$ 0.1255 \\
        \rowcolor{gray!15} \textit{vs.} 2nd Best & \boldgreen{+11.72\% $\uparrow$} & \boldred{--30.61\% $\downarrow$} \\
        \bottomrule
    \end{tabular}
    \label{tab:inference_time}
\end{table}

\begin{figure*}[t]    
    \centering
    \vspace{4pt}
    \includegraphics[width=1.00\linewidth]{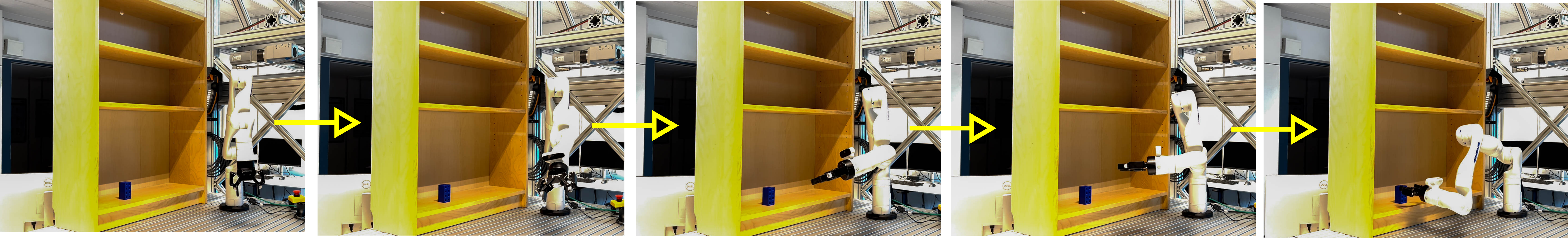}
    \vspace{-14pt}
    \caption{\textbf{Real-Robot Demonstration:} Using $\mathrm{X}$\textit{FlowMP}, the Kinova Gen3 manipulator is able to generate smooth, collision-free, and dynamically-feasible motions to reach the LEGO block on the shelf, given the positions of the robot and the target object as the task.}
    \vspace{-17pt}
    \label{fig:xflowmp_robot_exps}
\end{figure*}

\subsubsection{Inference Time}
Table \ref{tab:inference_time} benchmarks the inference time of $\mathrm{X}$\textit{FlowMP} against prior flow matching-based methods for both short- and long-horizon motion generation. While not the overall fastest, $\mathrm{X}$\textit{FlowMP} achieves the inference times of $0.1689$ and $0.2577$ seconds on the \texttt{RobotPointMass} and \texttt{LASA Handwriting} datasets, respectively. Notably, $\mathrm{X}$\textit{FlowMP} yields the planning/inference time of about $11.72\%$ faster than Albergo \textit{et al.}, which achieves second-best performance quantitatively in \texttt{RobotPointMass}, and about $0.06$ seconds slower than Tong \textit{et al.} that retains second-best performance quantitatively in \texttt{LASA Handwriting} dataset. Together with Table \ref{tab:comparison_results}, where $\mathrm{X}$\textit{FlowMP} demonstrates its gains in distributional alignment, generated trajectory quality, and energy efficiency, we acknowledge that these only lead to a modest increase in computational cost for long-horizon planning.

\subsection{Qualitative Results \& Limitations}
\subsubsection{Qualitative Results}
As shown in Fig. \ref{fig:xflowmp_quality}, for answering \textbf{Q5}, $\mathrm{X}$\textit{FlowMP} generates smooth and collision-free trajectories that follow expert demonstrations in the \texttt{RobotPointMass} environment (Fig. \ref{fig:xflowmp_quality}a). Meanwhile, other methods often encounter discontinuous and obstacle-colliding motions. In Fig. \ref{fig:xflowmp_quality}b, on the \texttt{LASA Handwriting} dataset, our method accurately captures the geometric structure of motion patterns, as illustrated in \texttt{Line}, \texttt{CShape}, \texttt{RShape}, \texttt{Angle}, and \texttt{JShape}, producing motions that remain well-aligned with expert references. In contrast, other approaches experience shape distortion and fail to maintain smooth curvature. Overall, we show that $\mathrm{X}$\textit{FlowMP} is able to demonstrate its strong contextualization and good quality across short- and long-horizon planning, producing qualitatively superior motions compared to all baseline methods. 

\subsubsection{Limitations}
 While our method scales well and produces smooth, task-adaptive motions, we find that its effectiveness largely depends on the task-agnosticity of the expert demonstrations. For example, failure cases occur when multiple experts propose outrageously different motions for a single task, resulting in collapsed learning distributions, even with a well-trained task-conditioned motion field.

\subsection{Real-Robot Experiments}
We validate the feasibility of $\mathrm{X}$\textit{FlowMP} on a real-robot system using the Kinova Gen3 manipulator. In this experiment, the task is defined by the positions of the robot and the target LEGO block placed on a shelf. $\mathrm{X}$\textit{FlowMP} is trained on $10$ expert demonstrations per planning mode, capturing smooth and dynamically-feasible trajectories under varying start–goal configurations. After training, the learned motion field is then deployed on the Kinova Gen3, connected to an
Intel Core i9-14900 CPU and an NVIDIA RTX 4080S
GPU. The robot is then able to generate collision-free motions to reach the LEGO block from different initial poses, given that the position of the LEGO block is known. As shown in Fig. \ref{fig:xflowmp_robot_exps}, the robot successfully executes the generated trajectory to reach the LEGO block while maintaining the smoothness and collision-free nature of expert demonstrations. Our demonstration videos are available in the supplementary materials.

%% file: 05_conclusions.tex
\section{Conclusions}
This paper presents $\mathrm{X}$\textit{FlowMP}, a task-conditioned generative motion planner that models trajectories as entropic flows via Schr{\"o}dinger bridges with score-based motion fields. Our method is able to generate smooth, collision-free, and dynamically feasible trajectories with high-order dynamics under environmental constraints through conditioning on start-goal configurations, enabling integration of task context into low-level motion generation. Through evaluations on the \texttt{RobotPointMass} and \texttt{LASA Handwriting} benchmarks, we show that $\mathrm{X}$\textit{FlowMP} achieves lower MMD, smoother motions, and lower energy consumption, while reducing short-horizon planning time and guaranteeing planning feasibility across both short- and long-horizon motions. Our real-world experiments on the Kinova Gen3 manipulator further confirm that the framework can reliably generate feasible trajectories, highlighting its potential for scalable, adaptable motion planning in real-world settings.